\documentclass[11pt]{article} 
\usepackage{rldmsubmit,palatino}
\usepackage{graphicx}
\usepackage{wrapfig}

\title{New Reinforcement Learning Using a Chaotic Neural Network\\ for Emergence of ``Thinking''\\
--- ``Exploration'' Grows into ``Thinking'' through Learning ---}

\author{
Katsunari ~Shibata\thanks{http://shws.cc.oita-u.ac.jp/\~{}shibata/home.html}  \ \& Yuki Goto\\
Department of Innovative (Electrical and Electronic) Engineering\\
Oita University\\
700 Dannoharu, Oita 870-1192, JAPAN \\
\texttt{katsunarishibata@gmail.com} \\
}

\begin{document}

\maketitle

\begin{abstract}
Expectation for the emergence of higher functions is getting larger
in the framework of end-to-end comprehensive reinforcement learning
using a recurrent neural network.
However, the emergence of ``thinking'' that is a typical higher function is difficult to realize 
because ``thinking'' needs non fixed-point, flow-type attractors
with both convergence and transition dynamics.
Furthermore, in order to introduce ``inspiration'' or ``discovery'' in ``thinking'',
not completely random but unexpected transition should be also required.

By analogy to ``chaotic itinerancy'', we have hypothesized  that {\bf ``exploration''
grows into ``thinking'' through learning by forming flow-type attractors
on chaotic random-like dynamics}.
It is expected that if rational dynamics are learned in a chaotic neural network (ChNN),
coexistence of rational state transition, inspiration-like state transition and
also random-like exploration for unknown situation can be realized.

Based on the above idea, we have proposed
{\bf new reinforcement learning using a ChNN}.
The positioning of exploration is completely different from the conventional one.
Here, {\bf the chaotic dynamics inside the ChNN produces exploration factors by itself}.
Since external random numbers for stochastic action selection are not used,
exploration factors cannot be isolated from the output.
Therefore, the learning method is also completely different from the conventional one.

One variable named causality trace is put at each connection,
and takes in and maintains the input through the connection according to the change in its output.
Using these causality traces and TD error,
the connection weights except for the feedback connections
are updated in the actor ChNN.

In this paper, as the result of a recent simple task to see whether the new learning works appropriately or not,
it is shown that a robot with two wheels and two visual sensors reaches a target
while avoiding an obstacle after learning
though there are still many rooms for improvement. 
\end{abstract}

\keywords{
\hspace{-6.5mm}thinking, chaotic neural network (ChNN), exploration, reinforcement learning (RL), function emergence
}

\acknowledgements{This research has been supported by JSPS KAKENHI
Grant Numbers JP23500245, JP15K00360
and many our group members}

\startmain 

\section{Introduction}
\vspace{-2mm}
Expectation for the emergence of artificial intelligence is growing these days
triggered by the recent results in reinforcement learning (RL) using a deep neural network (NN)\cite{DQN,AlphaGo}.
Our group has propounded for around 20 years that 
end-to-end RL from sensors to motors using a recurrent NN (RNN)
plays an important role for the emergence\cite{Intech,RLDM17}.
Especially, different from ``recognition'' whose inputs are given as sensor signals or 
``control'' whose outputs are given as motor commands,
higher functions are very difficult to design by human hands,
and the function emergence approach through end-to-end RL is highly expected.
Our group has shown that not only recognition and motion, but also
memory, prediction, individuality, and also similar activities to those in monkey brain at tool use
emerge\cite{RLDM17}.
We have also shown that a variety of communications emerge in the same framework\cite{RLDM17COM}.
However, the emergence of what can be called ``thinking'' that is one of the typical higher functions
has not been shown yet.
In this paper, the difficulty of emergence of ``thinking'' is discussed at first.
Then our hypothesis that ``exploration'' grows into ``thinking'' through learning
is introduced\cite{IJCNN15}.
To realize the hypothesis,
the use of a chaotic NN (ChNN) in RL
and new deterministic RL for it are introduced\cite{IJCNN15}.
Finally, it is shown that the new RL works in a simple task\cite{JCSS}
though that cannot be called ``thinking'' yet and there are still many rooms for improvement.
No other works with a similar direction to ours have not been found.

\section{Difficulty in Emergence of ``Thinking''}
\vspace{-2mm}
The definition of ``Thinking'' must be varied depending on the person.
However, we can ``think'' even when we close our eyes and ears, and
what we think does not change randomly, but logically or rationally.
Therefore, we hope many ones can agree that in order to realize ``thinking'',
rational multi-stage or flow-type state transition should be formed.

As a kind of dynamic functions, we have shown that a variety of memory-required functions
emerge in an RNN in simple tasks\cite{RLDM17}.
It is not so difficult to form memories as fixed-point convergence dynamics
if the initial feedback connection weights are set such that the transition matrix
for the connection is the identity matrix or close to it
when being linearly approximated.
That can also solve the vanishing gradient problem in error back propagation.
However, state transition needs not only convergence dynamics for association,
but also transition dynamics from one state to another.

Then, we employed a multi-room task in which
an agent moves and when it pushes a button,
one of the doors opens and then the agent can move to another room.
The sensor signals are the inputs of the agent's RNN, and
the RNN was trained based on RL from a reward when it reached the goal.
We expected that the internal state changed drastically between before and after the door open
though the difference in the sensor signals was not so large.
After learning, a large change in the internal state could be observed in some degree,
but the learning was very difficult\cite{Sawatsubashi}.

Furthermore, when we ``think'', ``inspiration'' or ``discovery'', which is a kind of unexpected
but not completely random and rational transition, must be also essential.
The convergence and transition dynamics seem contradict at a glance,
and it seems very difficult to form both dynamics from scratch in a regular RNN.
\begin{figure}[b]
\center
\begin{tabular}{ccc}
\begin{minipage}{.35\textwidth}
  \centering
  \includegraphics[height=3.4cm]{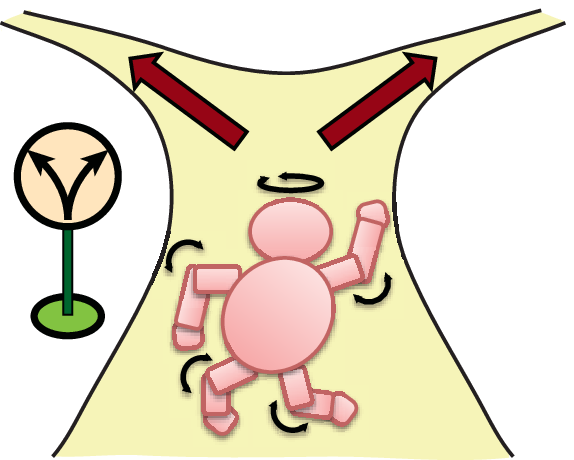}
  \caption{Lower or higher exploration\\ at a fork.}
  \label{fig:Fork}
\end{minipage}
\begin{minipage}{.50\textwidth}
  \centering
  \includegraphics[height=3.7cm]{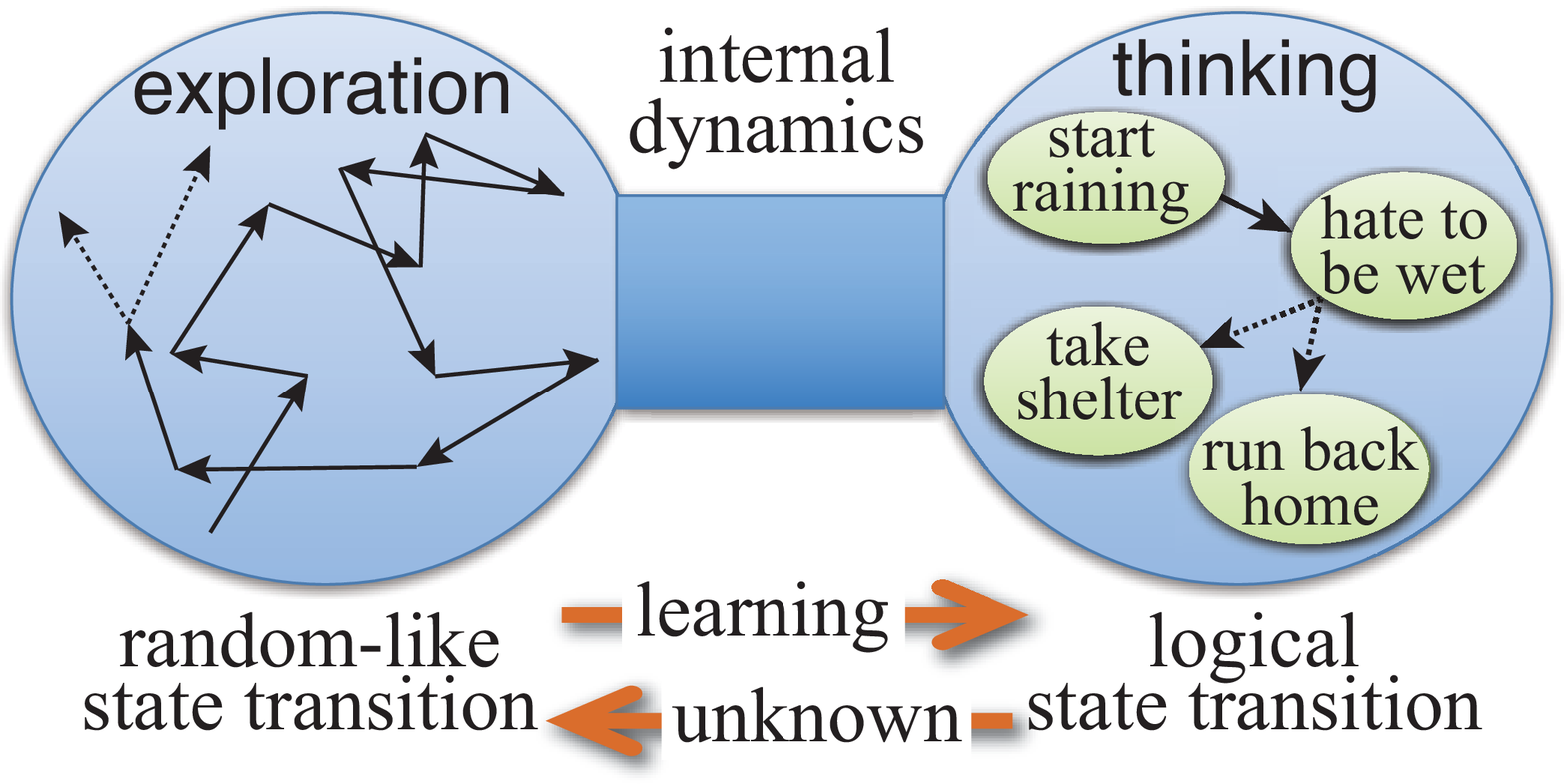}
  \caption{``Exploration'' and ``thinking'' are both a kind of internal dynamics
  and deeply related with each other.}
  \label{fig:ExplorationThinking}
\end{minipage}
\end{tabular}
\end{figure}

\section{Chaos and Hypothesis: Growth from ``Exploration'' to ``Thinking'' through Learning}
\vspace{-2mm}
Suppose we are standing at a forked road as shown in Fig. \ref{fig:Fork}.
Usually, we choose one from two options: going right and going left.
We do not care many other possible actions such as going straight and dancing.
It is not motor(actuator)-level lower exploration,
but higher exploration supported by some prior or learned knowledge\cite{Higher}.
Furthermore, at a fork, we may wonder such that
``this path looks more rough, but that way looks to go away from the destination''.
This can be considered as a kind of ``exploration'' and also as a kind of ``thinking''.
The place where the wave in mind occurs is inside the process before making a decision,
and learning should be reflected largely on the exploration.
The author's group has thought that exploration should be generated
inside of a recurrent NN (RNN) that generates motion commands\cite{Exploration}.
``Exploration'' and ``thinking'' are both generated as internal dynamics
as shown in Fig. \ref{fig:ExplorationThinking}.
``Exploration'' is more random-like state transitions.
On the other hand, ``thinking'' is more rational or logical state transition, and
sometimes higher exploration or unexpected but rational state transition
such as inspiration or discovery occurs in it.
\begin{figure}
  \centering
  \includegraphics[height=6.9cm]{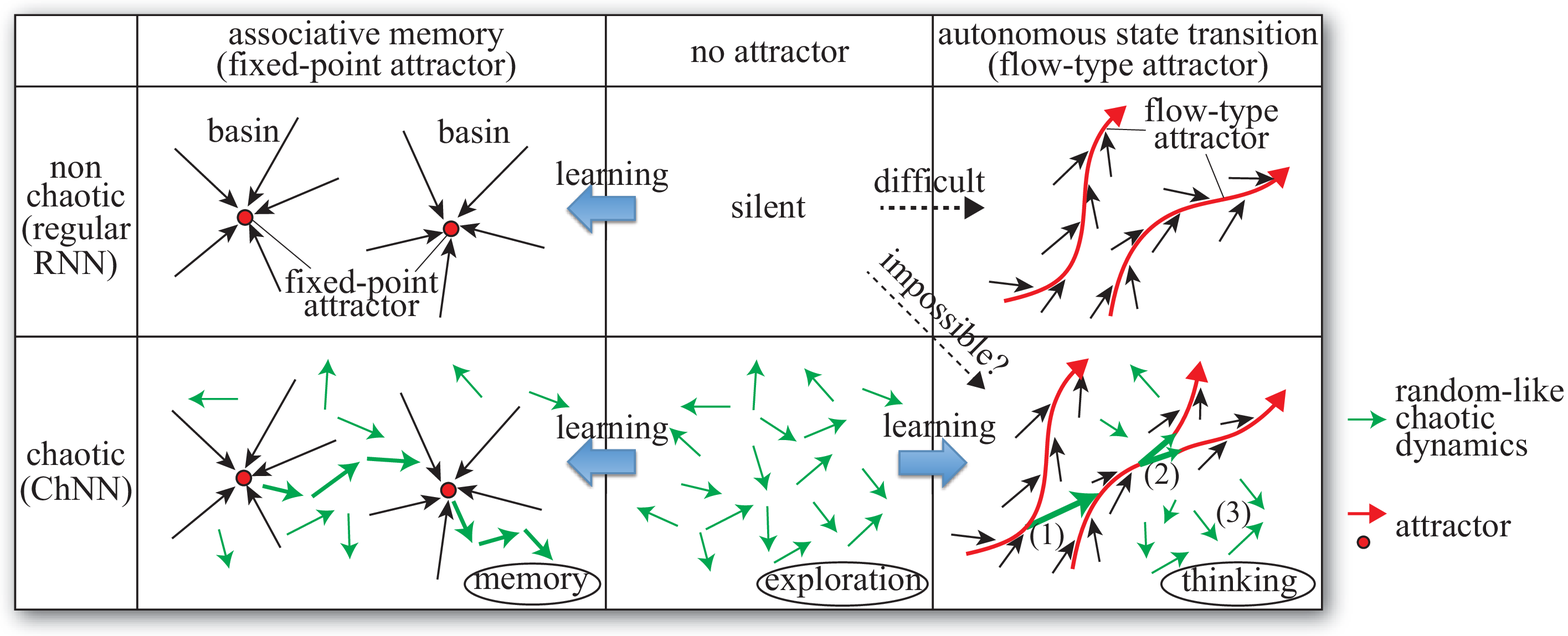}
  \vspace{-2mm}
  \caption{Rough schematic diagram of the combination of attractors and chaotic dynamics. See text in detail.}
  \label{fig:AttractorsAndChaos}
\end{figure}

Analogy to such dynamics can be found in chaotic dynamics.
In regular associative memory, fixed-point attractor (basin) is formed
around each memorized or learned pattern as shown in the upper left part in Fig. \ref{fig:AttractorsAndChaos}.
However, when a ChNN is used in it,
transition dynamics among the memorized patters called ``Chaotic Itinerancy''\cite{Itinerancy}
can be seen as the green arrows in the lower left part in Fig. \ref{fig:AttractorsAndChaos}.
If rational or logical state transitions are learned, it is expected that flow-type attractors are formed
as the red or black arrows in the lower right part in Fig. \ref{fig:AttractorsAndChaos}.
It is also expected that as the green arrows,
(1)inspiration or discovery emerges as irregular transitions among the attractors but reflecting the distance,
(2)higher exploration emerges at branches of the flow, 
and (3)in unknown situations, no attractor is formed, and remaining random-like chaotic dynamics appears
until return to a known situation.
Skarda et al. reported that the activities on the olfactory bulb in rabbits become
chaotic for unknown stimuli\cite{Skarda}.
Osana et al. have shown the difference of chaotic property between known and unknown
patterns on an associative memory using a ChNN,
and also after an unknown pattern is learned,
association to the pattern is formed as well as the other known patterns\cite{Osana}.
From the above discussion, we have hypothesized  that {\bf ``exploration''
grows into ``thinking'' through learning by forming flow-type attractors
on chaotic random-like dynamics and that can be realized on reinforcement learning using a ChNN}.

\section{New Reinforcement Learning (RL) Using a Chaotic Neural Network (ChNN)}
In order to realize the above idea, our group has proposed
new reinforcement learning using a ChNN\cite{IJCNN15}.
The positioning of exploration in learning is completely different from the conventional one.
Here, the chaotic dynamics inside the ChNN produces exploration factors by itself.
Since external random numbers for stochastic action selection are not used,
exploration factors cannot be isolated from the output.
Then the learning method has to be completely different from the conventional one.

\begin{figure}
\centering
\includegraphics[height=6.4cm]{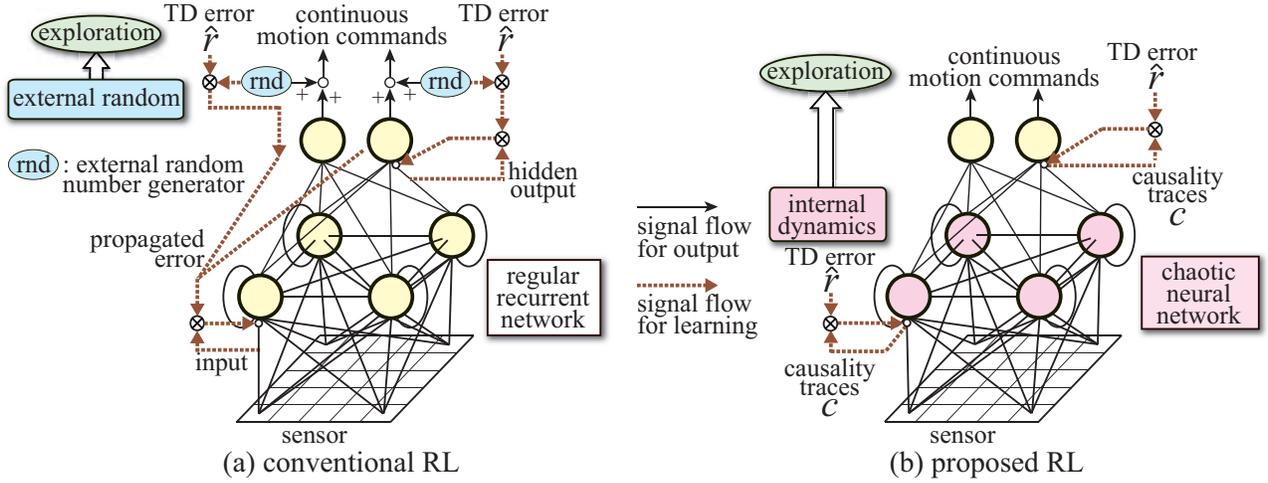}
\caption{Comparison of conventional RL and proposed RL\\ (only actor network)
\cite{IJCNN15}}
\label{fig:ChaosNN}
\end{figure}

Assuming that the motions are continuous, actor-critic type reinforcement learning architecture
is employed.
Here, to isolate the chaotic dynamics from the critic,
the actor is implemented in a ChNN and the critic is implemented in another regular
layered NN as shown in Fig. \ref{fig:Task}.
The inputs are the sensor signals for the both networks.
Here, only the learning of actor ChNN that is largely different from the conventional
reinforcement learning is explained comparing with the conventional one using Fig. \ref{fig:ChaosNN}.
In our conventional works, as shown in Fig. \ref{fig:ChaosNN}(a),
by adding a random number (noise) $rnd_{j,t}$ to each actor output, the agent explores.
The actor network is trained by BPTT (Back Propagation Through Time)
using the product of the random number and TD error as the error for each output of the RNN.

In the proposed method, there is no external random numbers added to the actor outputs.
The network is a kind of RNN, but by setting each feedback connection to a large random value,
it can produce chaotic dynamics internally.
Because the learning of recurrent connections does not work well,
only the connections from inputs to hidden neurons and from hidden neurons to output neurons
are trained.
One variable $C_{ji}$ named causality trace is put on each connection, and
takes in and maintains the input through the connection according to the change in its output as
\begin{equation}
C^{[l]}_{ji,t}=(1-|\Delta x^{[l]}_{j,t}|)C^{[l]}_{ji,t-1}+\Delta x^{[l]}_{j,t} x^{[l-1]}_{i,t}
\label{Eq:C-Trace-RL}
\end{equation}
where $x^{[l]}_{j,t}$: output of the $j$-th neuron in the $l$-th  layer at time $t$,
$\Delta x_t = x_t - x_{t-1}$.\vspace{1mm}\\
Using the causality trace $C_{ji}$ and TD error ${\hat r}_t$,
the weight $w^{[l]}_{ji}$ from the $i$-th neuron in $(l\!\!-\!\!1\!)$-th layer
to the $j$-th neuron in the $l$-th layer is updated with a learning rate $\eta$ as
\begin{equation}
\vspace{-1mm}
\Delta w^{[l]}_{ji,t}=\eta {\hat r_t} C^{[l]}_{ji,t}.
\label{Eq:LearnActor}
\end{equation}

\section{Learning of Obstacle Avoidance}
\begin{wrapfigure}[26]{r}{80mm}
\vspace*{-\intextsep}
     \centering
       \includegraphics[height=8.4cm]{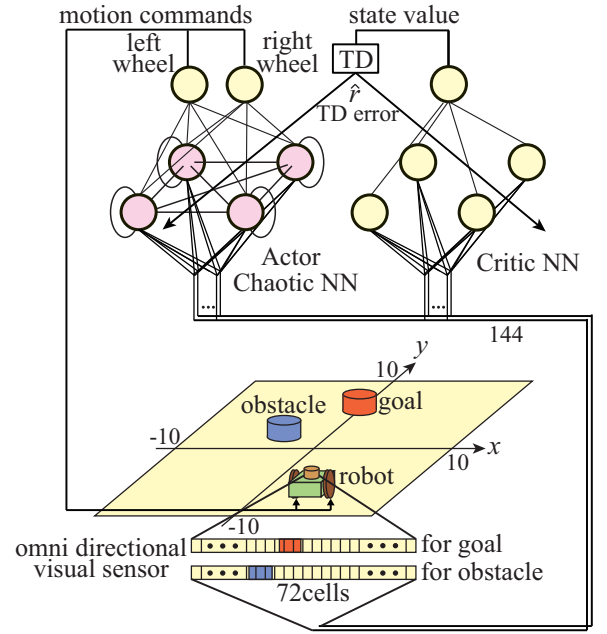}
       \caption{Learning of obstacle avoidance for a robot with two wheels and two sensors using a ChNN.}
       \label{fig:Task}
\end{wrapfigure}
Since the learning is completely different from the conventional one,
it is necessary to show whether the learning works appropriately in a variety of tasks or not.
It was already applied to several tasks\cite{IJCNN15}\cite{Higher}.
In this paper, the result of a recent task in which a robot has two wheels
and two visual sensors is shown\cite{JCSS}.
As shown in Fig. \ref{fig:Task}, there is a $20 \!\!\times \!\!20$ size of field,
and its center is the origin of the field.
What the robot has to do is to reach a goal while avoiding an obstacle.
The goal is put on the location (0, 8) with radius $r\!\!=\!\!1.0$, and a robot ($r\!\!=\!\!0.5$) and
an obstacle ($r\!\!=\!\!1.5$) are put randomly at each episode. 
The orientation of the robot is also decided randomly.
Each of the two omnidirectional visual sensors has 72 cells, and catches only goal or obstacle respectively.
Total of 144 sensor signals are the inputs of both critic and actor networks, and
the right and left wheels rotate according to the two actor outputs.
When the robot comes in the goal area, a reward is given, and when it collides
with the obstacle, a small penalty is given.
One episode is defined as until arrival to the goal or 1000 steps from start. 
The both NNs have three layers including input layer.
The number of hidden neurons is 10 for critic network and 100 for actor ChNN.
\begin{figure*}[]
\centering
\includegraphics[scale=0.71]{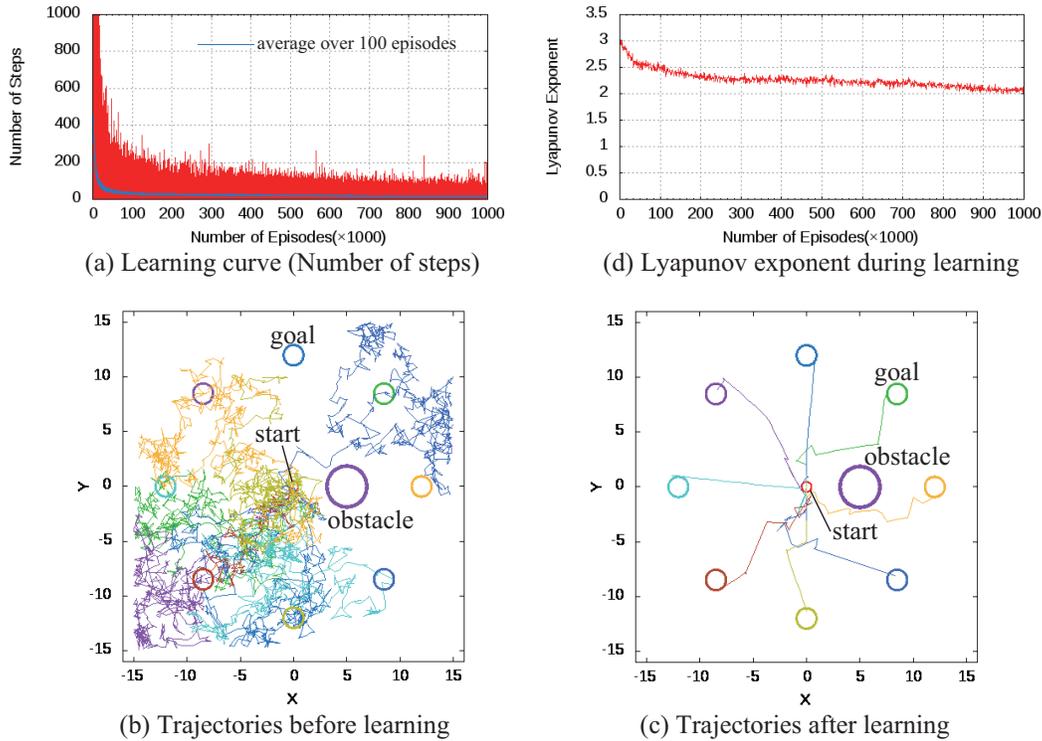}
\vspace{-1mm}
\caption{Learning results\cite{JCSS}. In (b) and (c), the initial robot location is on the center and
its initial orientation is for upper-side in the figure. Actually, the goal is located at the same place,
but on this figure the location is varied relatively.}
\label{fig:Results}
\end{figure*}

The learning results are shown in Fig. \ref{fig:Results}.
Fig. \ref{fig:Results}(a) shows learning curve and the vertical axis indicates
the number of steps to the goal.
The blue line shows its average over each 100 episodes.
It can be seen that according to the number of episodes,
the number of steps decreases.
Fig. \ref{fig:Results}(b) and (c) show the robot trajectories before and after learning respectively.
Actually, the goal location is fixed and initial robot location is varied.
But to see the initial robot orientation, the robot is on the center of the field,
and its orientation is for upper-side initially on the figure.
Relatively, the goal location is varied.
The size of the obstacle in the figure shows the area where the robot
collides with the obstacle and then is larger than the actual size.
Before learning, the robot explored almost randomly, but after learning,
it could reach the goal while avoiding the obstacle.
In conventional learning, randomness of action selection is often decreased
as learning progresses, but here, it can be seen that exploration factors
decreased autonomously according to learning.
However, there are some strange behaviors such that the robot changes its direction
suddenly.
It was observed that the actor hidden neurons were likely to have a value around the maximum 0.5
or minimum -0.5.
There is still a large space to improve the learning method.

Fig. \ref{fig:Results}(d) shows Lyapunov exponent to see the chaotic property of this system
including the environment.
The robot is located at one of the 8 locations in Fig. \ref{fig:Results}(c), 
and the obstacle is also located at one of 8 locations.
For each of the $8 \times 8 = 64$ combinations,
a small perturbation vector with the size $d_{before}$ is added to the internal states of hidden neurons,
and the distance $d_{after}$ \hspace{-0.1mm} in the internal states at the next time
between the cases of no perturbation and addition of the perturbation
is compared.
The average of $ln(d_{after}/d_{before})$ is used as the exponent here.
From the figure, it can be seen that Lyapunov exponent is gradually
decreased, but since the value is more than 0.0,
it is considered that the chaotic property is maintained.
In \cite{IJCNN15}, it was observed that when the environment changed,
Lyapunov exponent increased again.

\end{document}